\documentclass[10pt,twocolumn,letterpaper]{article}
\usepackage[pagenumbers]{cvpr}
\usepackage[T1]{fontenc}
\usepackage{microtype,booktabs}
\usepackage{amsmath,amssymb,mathtools}
\usepackage{array,multirow,xspace,tabularx}
\newcommand{\method}{PreWAM\xspace}

\usepackage{xcolor,colortbl}
\definecolor{TableHeader}{HTML}{EAF3F8}
\definecolor{TableOurs}{HTML}{E5F2F7}

\usepackage{placeins}
\definecolor{cvprblue}{rgb}{0.21,0.49,0.74}
\usepackage[pagebackref,breaklinks,colorlinks,allcolors=cvprblue]{hyperref}
\def\confName{CVPR}
\def\confYear{2027}
\title{PreWAM: Exploring the Design of Latent Future Prediction for World-Action Modeling}
\author{
\textbf{Wan Chi}$^{1,2}$\quad \textbf{Sen Li}$^{2,\dagger}$\quad \textbf{Zihang Cui}$^{2,3}$\quad \textbf{Chuqin Huang}$^{2,4}$\\
\textbf{Yunlong Wang}$^{2}$\quad \textbf{Nanshan Deng}$^{2,5}$\quad \textbf{Yanjun Huang}$^{1}$\\[6pt]
{\normalsize $^{1}$Tongji University\quad $^{2}$Rimbot}\\
{\normalsize $^{3}$Xi'an University of Electronic Science and Technology}\\
{\normalsize $^{4}$Southeast University\quad $^{5}$Tsinghua University}
}
\begin{document}
\maketitle
\begingroup
\renewcommand{\thefootnote}{\fnsymbol{footnote}}
\footnotetext[2]{Corresponding author.}\footnotetext[0]{\href{mailto:2532900@tongji.edu.cn}{2532900@tongji.edu.cn}}
\endgroup
\begin{abstract}
World-action models combine action generation with future prediction, allowing visual transitions to serve as supervision for policy learning. Our controlled study of prediction spaces shows that spatially structured semantic features support better overall generalization under visual perturbations and make motion information more readily recoverable from learned predictive representations. Building on these observations, we introduce PreWAM, a world-action model that organizes future supervision around state changes in a prediction space defined by pretrained DINO features. Specifically, Residual World Prediction jointly learns source-relative changes and their resulting future states to emphasize dynamic information while preserving future-state semantics. Temporal-Scale Prediction conditions world queries on temporal distance to capture both short-term local motion and longer-horizon state evolution. These predictive objectives jointly shape a shared vision-language backbone with action supervision during pretraining and downstream policy learning. Experiments across the LIBERO benchmark family, SimplerEnv, RoboTwin 2.0, and real-world robotic manipulation demonstrate strong manipulation performance and improved generalization and robustness under diverse distribution shifts.
\href{https://newbrains1.github.io/PreWAM/}{Project page}
\end{abstract}

\section{Introduction}\label{sec:intro}
Robot demonstrations specify what action to take, while the accompanying observations reveal how the scene changes as that action unfolds. World-action modeling uses both signals, combining action learning with prediction of future observations. Latent prediction makes this combination particularly attractive: a visual encoder supplies dense supervision from videos, and a policy can learn to anticipate transitions without reconstructing every pixel \citep{vlajepa,jepawam}.

Making future supervision useful for control involves two design dimensions: the representation of future observations and the formulation of their prediction. Future frames mix persistent content, manipulation-induced changes, and appearance variations, whose relative emphasis depends on the prediction space. Image self-supervision, video reconstruction, and predictive video learning impart different biases toward spatial correspondence, appearance detail, and temporal structure. We investigate how these representational biases affect the information a policy learns from future supervision and its ability to act beyond the training distribution.

Our controlled study reveals that prediction spaces with similar standard-task performance differ in their robustness to visual changes. Spatial semantic supervision provides the strongest overall generalization, while reconstruction supervision retains advantages in some categories. Displacement readout provides complementary evidence: motion information remains more accessible under perturbations in policies trained with spatial semantic targets.

The second dimension concerns how prediction targets are organized over time. Supervising only the future state leaves the change from the source implicit, while a fixed prediction interval restricts supervision to a single temporal scale. We introduce \emph{Residual World Prediction} (RWP) and \emph{Temporal-Scale Prediction} (TSP) to make both relationships explicit. RWP predicts changes relative to a source state and jointly supervises these changes and the future states they reconstruct. TSP conditions world queries on temporal distance, providing supervision over several horizons within the same clip.

We instantiate this design in \method, where world prediction and action learning jointly shape a shared vision-language backbone. Pretraining combines robot demonstrations, which provide both objectives, with human videos that supply additional predictive supervision. Downstream posttraining continues joint prediction and action learning on the target manipulation data. At deployment, only the policy backbone and action expert are required.

We evaluate PreWAM on standard manipulation tasks and a range of distribution shifts. It achieves 98.1\% success on LIBERO, 79.1\% on LIBERO-Plus, and 58.7\% on LIBERO-Pro. LIBERO-Pro2.0 further tests robustness to static and runtime disturbances. SimplerEnv and RoboTwin 2.0 further extend the evaluation to real-to-sim transfer and multitask bimanual manipulation. Three tasks on a physical Piper platform further evaluate single-arm and bimanual manipulation. Our contributions are:
\begin{itemize}
\item Controlled comparisons reveal that prediction spaces supporting similar in-distribution performance can yield substantially different policy generalization, with spatial semantic supervision providing stronger overall robustness to visual perturbations.
\item We introduce PreWAM, a world-action model that organizes future supervision around state changes across temporal horizons, allowing predictive learning to shape the shared representations used for action generation.
\item Experiments demonstrate strong generalization across the LIBERO benchmark family and effective manipulation in SimplerEnv and RoboTwin 2.0, complemented by real-world evaluation of single-arm and bimanual tasks on a Piper platform.
\end{itemize}

% Queue the overview figures after the introduction for pages 3 and 4.
\begin{figure*}[t]
\centering
\includegraphics[width=\textwidth]{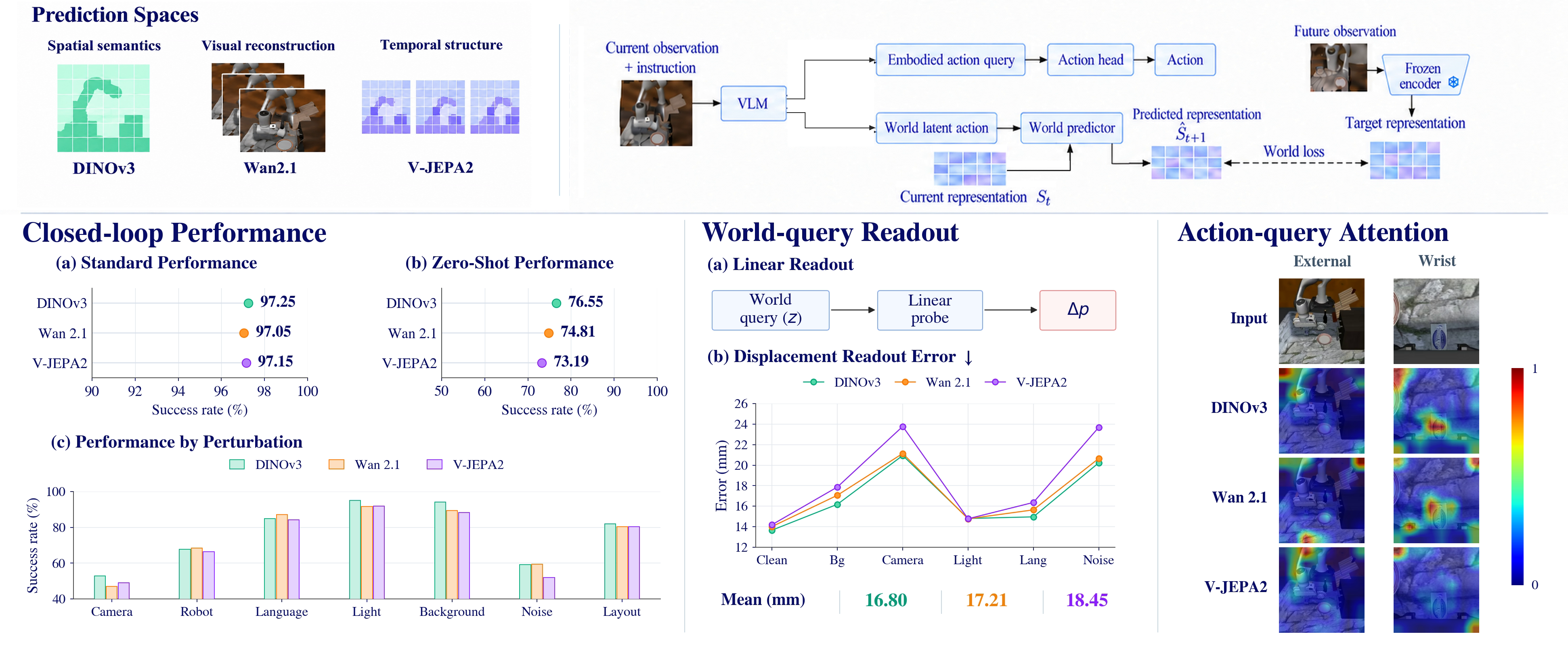}
\caption{\textbf{World representations and policy generalization.} Left: success on LIBERO and LIBERO-Plus. Center: end-effector displacement readout from world latent actions. Right: attention maps from embodied action queries on the same observations.}
    \label{fig:representations}
\end{figure*}

\begin{figure*}[t]
\centering
\includegraphics[width=\textwidth,trim=0 0 0 70,clip]{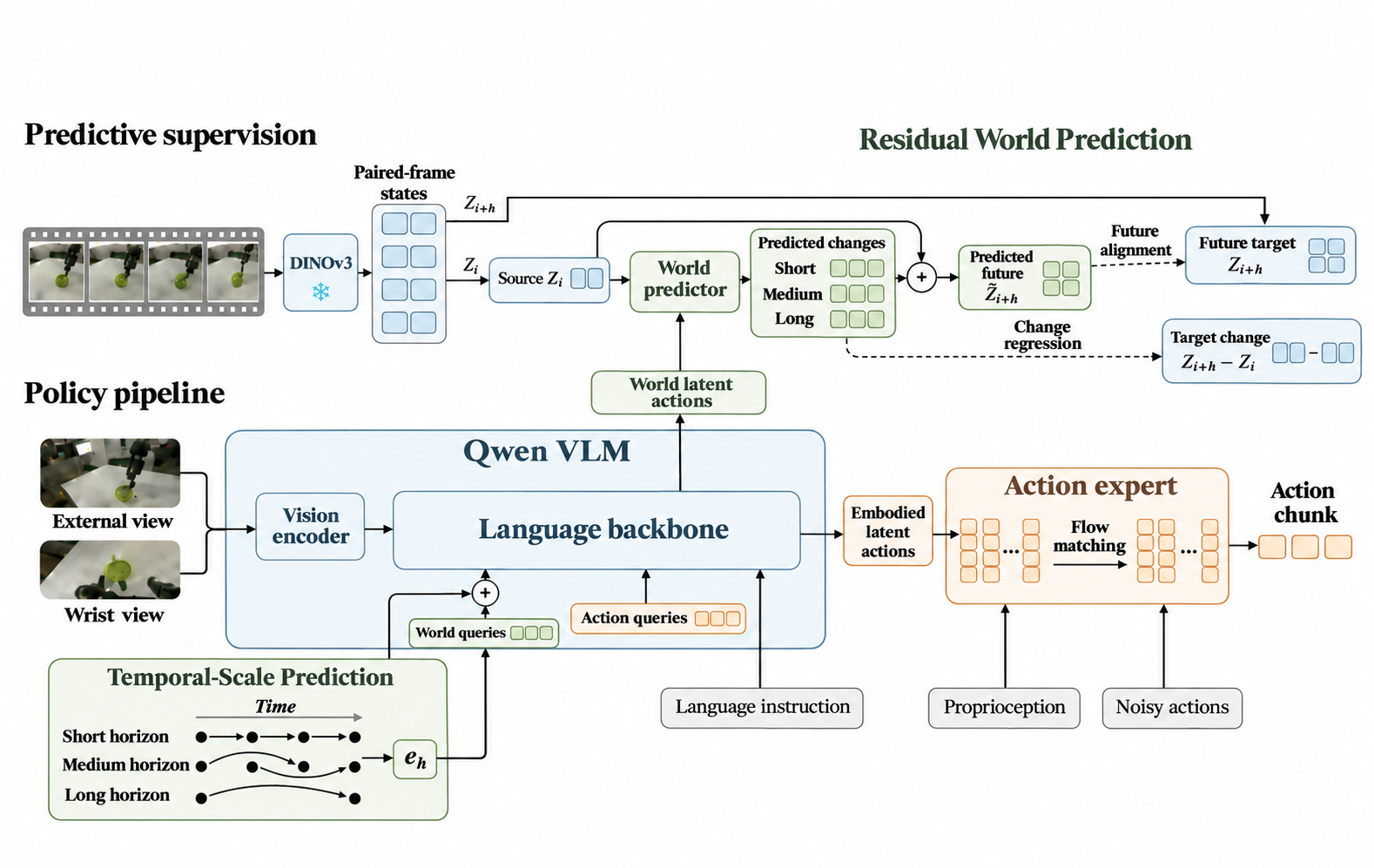}
\caption{\textbf{PreWAM overview.} A shared Qwen VLM processes current observations and language instructions to provide features for the action expert and world predictor. The predictive branch uses paired-frame features from frozen DINOv3 as visual states, learning source-to-future changes through Residual World Prediction across short, medium, and long horizons through Temporal-Scale Prediction. Joint supervision of changes and future states trains the VLM alongside action learning; deployment uses only the VLM and action expert.}
\label{fig:framework}
\end{figure*}

\section{Related Work}\label{sec:related}
\textbf{Vision-Language-Action Models.}
Vision-language-action models transfer the perceptual and linguistic knowledge of pretrained multimodal backbones to robot control. Action-token prediction provides a direct extension of vision-language modeling to robot demonstrations \citep{openvla}, while dedicated continuous-action experts generate action sequences from multimodal representations \citep{pi05,gr00t}. These methods learn to map observations and language instructions to actions from robot demonstrations, making the quality of multimodal representations an important foundation for policy learning.

\textbf{Representations for Prediction and Control.}
Visual pretraining objectives emphasize different aspects of an observation. Reconstruction preserves appearance information, image self-supervision learns dense semantic features, and predictive video learning captures temporal regularities in representation space \citep{wan,dinov3,vjepa2}. When these representations become prediction targets, their properties also influence the information emphasized during policy training. Feature-space world models use pretrained visual representations for future-state prediction and action-conditioned planning without reconstructing pixels \citep{dinoworld}.

\textbf{World Modeling for Policy Learning.}
World-action models couple visual prediction with robot control, but differ in how predictive information reaches the policy. Video-based architectures can connect world and action experts through shared latent interfaces, allowing joint supervision to shape the visual features used for action generation \citep{bridgewam}. Prediction can also operate directly in a pretrained feature space without reconstructing future images. Such methods predict future features or jointly model current and future representations to provide auxiliary supervision for action learning \citep{vlajepa,jepawam}, and can also learn from videos without action annotations \citep{vlajepa}. PreWAM explicitly models changes relative to source states and organizes predictive supervision across multiple temporal horizons.

\section{World Representations for Policy Generalization}\label{sec:representations}
World representations define the visual content that future prediction brings into policy learning. How does the content emphasized by these representations affect a policy when observations depart from familiar scenes? We approach this question through policy evaluation under perturbations and motion readout from the learned predictive representations. We compare spatial semantic features from DINOv3 \citep{dinov3}, reconstruction latents from the Wan 2.1 video VAE \citep{wan}, and predictive video features from V-JEPA 2 \citep{vjepa2}. The comparison uses direct future-state prediction within a common Qwen-based policy framework and training data, with each frozen encoder supplying its own visual prediction targets. Downstream training uses the same optimization settings across prediction spaces, with a global batch size of 256 and 36,000 updates. 

\subsection{Generalization across Prediction Spaces}
Similar task success in familiar scenes masks differences in the effects of predictive supervision. The three policies fall within a 0.20-point range on LIBERO, yet separate more clearly under the seven perturbation categories of LIBERO-Plus (Figure~\ref{fig:representations}). Spatial semantic supervision provides the strongest aggregate performance, while the relative strengths of the other spaces depend on the perturbation. 

The category results help characterize these distinctions. Spatial semantic supervision has its clearest advantages over reconstruction supervision under camera, background, and lighting changes. Such changes can substantially alter an image while preserving the intended interaction, making correspondence across appearances useful for control. This pattern is consistent with the benefit of targets that preserve spatial scene content without emphasizing every appearance detail. Reconstruction supervision remains competitive: it leads on the Robot and Language categories and closely matches spatial semantic supervision under noise. Its Robot performance suggests that retaining fine visual configurations can also be useful when the robot's appearance changes. Overall, DINOv3 provides the most effective predictive supervision for generalization in this comparison.

\subsection{Motion Information in Policy Representations}
\textbf{Displacement readout from world queries.}
We first examine whether the world queries supplied to the world predictor contain motion information. We mean-pool their final-layer VLM features and fit a linear ridge probe to three-dimensional end-effector displacement. The probe is fitted on clean observations and held fixed across perturbations. We evaluate on held-out episodes using five-fold cross-validation grouped by episode. Across the three prediction spaces, mean readout errors range from 16.80 to 18.45 mm, well below the 55.10 mm baseline that always predicts the training-fold mean displacement, indicating that world queries encode recoverable motion information.

Readout errors are close on clean observations but diverge under camera shifts and noise. Spatial semantic and reconstruction supervision both outperform predictive video supervision in these conditions, with spatial semantic supervision yielding the lowest error, also under background changes. These results show that world queries learned with spatial semantic supervision retain more readily accessible motion information under these visual perturbations.

\textbf{Action-query attention.}
Figure~\ref{fig:representations} visualizes attention from the action queries that condition the action expert in a task that places cheese on a plate. With spatial semantic supervision, these queries focus on manipulation-relevant regions in both external and wrist views. Reconstruction and predictive video supervision show less consistent focus across views, with more diffuse responses or attention away from the manipulation region in one view. This example illustrates the task-relevant visual focus of action queries learned with spatial semantic supervision.

The study above shows that the prediction space affects policy generalization, the accessibility of motion information in world queries, and the visual attention of action queries. We next investigate how to organize prediction targets within this space, explicitly modeling source-relative changes and temporal distance through Residual World Prediction and Temporal-Scale Prediction.

\section{PreWAM}\label{sec:method}
\subsection{Overview}
\textbf{Shared policy architecture.}
PreWAM couples action generation and visual transition prediction through a shared vision-language backbone (Figure~\ref{fig:framework}). Given current multi-view observations $o_t$, a language instruction $\ell$, and available proprioception $s_t$, it predicts an action chunk $A_t=(a_t,\ldots,a_{t+K-1})$. Learned world queries $Q^{\mathrm{wm}}$ and embodied action queries $Q^{\mathrm{act}}$ are processed together with observation and language tokens:
\begin{equation}
(H_t^{\mathrm{wm}},H_t^{\mathrm{act}})
=F_\theta(o_t,\ell,\widetilde Q^{\mathrm{wm}},Q^{\mathrm{act}}).
\label{eq:backbone}
\end{equation}
Here $\widetilde Q^{\mathrm{wm}}$ includes the horizon information introduced in Section~\ref{sec:tsp}. The world predictor decodes world latent actions $H_t^{\mathrm{wm}}$ into visual transitions; the action expert decodes embodied latent actions $H_t^{\mathrm{act}}$ into robot actions.

\textbf{Action generation.}
The action expert uses the embodied latent actions $H_t^{\mathrm{act}}$ and available proprioception $s_t$ to generate an action chunk. Given a demonstration $A_t$ and Gaussian noise $\epsilon$, we form $A_t^\tau=(1-\tau)\epsilon+\tau A_t$ at flow time $\tau\in[0,1]$. The expert $v_\psi$ learns the velocity toward the demonstrated actions:
\begin{equation}
\mathcal L_{\mathrm{act}}=
\mathbb E_{A_t,\epsilon,\tau}\!\left[
\left\|v_\psi(A_t^\tau,\tau\mid H_t^{\mathrm{act}},s_t)
-(A_t-\epsilon)\right\|_2^2\right].
\label{eq:action}
\end{equation}

\textbf{Visual world states.}
The backbone processes the current observations and instruction; a frozen encoder extracts states from the training clip for the world predictor and its supervision. For view $v$, a visual encoder $E$ maps each sampled frame $x_j^v$ to a patch grid. Two consecutive sampled frames form one state by channel-wise concatenation:
\begin{equation}
Z_i^v=\operatorname{concat}_{\mathrm{ch}}
\big(E(x_{2i}^v),E(x_{2i+1}^v)\big).
\label{eq:paired}
\end{equation}
Each state retains both the spatial layout and the ordering of its two observations. Pairing makes a local temporal interval available in the target without pooling away patch structure. Across these pairs, the world objective learns transitions over longer intervals. We use four paired states per clip and preserve each available camera view as a separate prediction target.

\subsection{Residual World Prediction}
Much of a manipulation scene remains unchanged, while robot motion and object interactions appear as local changes. RWP provides the source state explicitly and predicts the feature change needed to reach a future state, focusing the prediction branch on how the scene evolves. For source index $i$, view $v$, and horizon $h$, the world predictor produces
\begin{align}
\widehat D_{i,h}^v&=P_\phi(Z_i^v,H_{i,h}^{\mathrm{wm}}),\\
\widehat Z_{i+h}^v&=Z_i^v+\widehat D_{i,h}^v.
\label{eq:residual}
\end{align}
The additive path preserves existing scene information, while the residual describes the change relative to it. We supervise both the reconstructed future state and the predicted change, retaining future scene semantics while learning its dynamics. Future alignment measures cosine distance at each spatial position:
\begin{equation}
\mathcal L_{\mathrm{future}}^{i,h,v}
=1-\frac{\langle\widehat Z_{i+h}^v,Z_{i+h}^v\rangle}
{\|\widehat Z_{i+h}^v\|_2\|Z_{i+h}^v\|_2}.
\end{equation}
Direct change supervision regresses the feature difference:
\begin{equation}
\mathcal L_{\mathrm{change}}^{i,h,v}
=\operatorname{SmoothL1}\big(\widehat D_{i,h}^v,Z_{i+h}^v-Z_i^v\big).
\label{eq:change}
\end{equation}
Future alignment constrains the semantic direction of the reconstructed state, while change regression constrains the magnitude of changes across feature channels, complementing the scale invariance of cosine alignment. All source and target states are extracted by the frozen encoder. Through the world queries, these objectives supervise the shared VLM, making state changes an explicit representation-learning signal alongside action learning.

\subsection{Temporal-Scale Prediction}\label{sec:tsp}
RWP describes what changes relative to a source state; TSP organizes these changes across temporal horizons. Shorter horizons capture local robot and object motion, while longer horizons capture the spatial configurations that emerge as these motions accumulate. TSP supervises transitions over short, medium, and long horizons among four paired states sampled across the training clip, teaching the shared representation both local motion and longer-range state evolution. Horizons measure relative distance between sampled states, with physical duration determined by the sampling interval. All query groups are processed together in a single VLM forward pass conditioned on the current observations and instruction. Each valid source--horizon pair receives a group of world queries, with a learned horizon embedding added before VLM processing:
\begin{equation}
\widetilde Q_{i,h}^{\mathrm{wm}}=Q_{i,h}^{\mathrm{wm}}+e_h.
\end{equation}
Conditioning queries on the horizon before VLM processing allows them to organize predictive information for different temporal ranges. A shared world predictor then decodes them into the corresponding residual transitions.

We first average within each horizon over source indices, available views, and spatial positions, then across the horizon set $\mathcal H$:
\begin{equation}
\mathcal L_{\mathrm{wm}}=
\frac{1}{|\mathcal H|}\sum_{h\in\mathcal H}
\left\langle\mathcal L_{\mathrm{future}}+
\beta\mathcal L_{\mathrm{change}}\right\rangle_h.
\label{eq:world}
\end{equation}

% Queue benchmark floats early so the real-world figure has its own page top.
\begin{table*}[!tp]
\centering
\caption{Success rates (\%) on LIBERO and zero-shot LIBERO-Plus. LIBERO-Plus reports success across seven perturbation categories.}
\label{tab:libero}
\label{tab:plus}
\footnotesize
\setlength{\tabcolsep}{4pt}
\begin{tabular}{@{}>{\raggedright\arraybackslash}p{0.235\textwidth}rrrrrrrrr@{}}
\toprule
\rowcolor{TableHeader}
 & LIBERO & \multicolumn{8}{c}{LIBERO-Plus} \\
\cmidrule(lr){2-2}\cmidrule(lr){3-10}
\rowcolor{TableHeader}
Method & Avg. & Camera & Robot & Language & Light & Background & Noise & Layout & Avg. \\
\midrule
OpenVLA \citep{openvla} & 76.5 & 0.8 & 3.5 & 23.0 & 8.1 & 34.8 & 15.2 & 28.5 & 16.3 \\
OpenVLA-OFT \citep{openvlaoft} & 97.1 & 56.4 & 31.9 & 79.5 & 88.7 & 93.3 & 75.8 & 74.2 & 71.4 \\
$\pi_0$ \citep{pi0} & 94.1 & 13.8 & 6.0 & 58.8 & 85.0 & 81.4 & 79.0 & 68.9 & 56.1 \\
UniVLA \citep{univla} & 95.2 & 1.8 & 46.2 & 69.6 & 69.0 & 81.0 & 21.2 & 31.9 & 45.8 \\
JEPA-WAM \citep{jepawam} & 96.7 & \textbf{79.2} & 59.2 & 68.2 & 93.3 & 94.6 & \textbf{83.6} & 76.1 & \textbf{79.2} \\
Fast-WAM \citep{fastwam} & 97.6 & 16.4 & 44.5 & 68.9 & 78.2 & 53.7 & 37.7 & 60.7 & 51.5 \\
Motus \citep{motus} & 97.7 & 17.9 & 43.3 & 74.1 & 54.7 & 44.9 & 26.2 & 52.0 & 44.7 \\
ResVLA \citep{resvla} & 96.6 & 49.8 & 59.9 & 88.5 & 90.5 & \textbf{94.9} & 76.8 & 79.0 & 77.1 \\
LaWAM \citep{lawam} & \textbf{98.6} & 37.2 & 71.6 & 87.1 & 93.5 & 94.1 & 69.4 & \textbf{83.3} & 76.6 \\
\midrule
\rowcolor{TableOurs}
\textbf{PreWAM} & 98.1 & 54.6 & \textbf{73.0} & \textbf{89.3} & \textbf{97.4} & 93.9 & 62.7 & 83.1 & 79.1 \\
\bottomrule
\end{tabular}
\end{table*}

\begin{table}[!tbp]
\centering
\caption{LIBERO-Pro success (\%). \textdagger: recomputed means.}
\label{tab:pro}
\footnotesize
\setlength{\tabcolsep}{2pt}
\begin{tabularx}{\columnwidth}{@{}l*{5}{>{\raggedleft\arraybackslash}X}@{}}
\toprule
\rowcolor{TableHeader}
Method & Spatial & Object & Goal & Long & Avg. \\
\midrule
OpenVLA \citep{openvla} & 56.6 & 39.2 & 58.4 & \textbf{52.4} & 51.7\textsuperscript{\textdagger} \\
$\pi_0$ \citep{pi0} & 50.4 & 42.6 & 45.2 & 37.6 & 44.0 \\
$\pi_{0.5}$ \citep{pi05} & 52.2 & 57.0 & 55.6 & 48.0 & 53.2\textsuperscript{\textdagger} \\
VLA-JEPA \citep{vlajepa} & 55.4 & 50.0 & 45.0 & 42.6 & 48.3 \\
Fast-WAM \citep{fastwam} & 57.6 & 49.2 & 58.1 & 40.8 & 51.4 \\
ImageWAM \citep{imagewam} & 63.0 & 47.6 & 42.2 & 42.0 & 48.7 \\
LaWAM \citep{lawam} & 63.2 & 54.6 & 46.0 & 47.0 & 52.7 \\
JEPA-WAM \citep{jepawam} & 48.2 & 43.2 & 37.6 & 38.6 & 41.9 \\
BridgeWAM \citep{bridgewam} & 57.8 & 55.1 & 58.4 & 43.5 & 53.7 \\
Faster-WAM \citep{fasterwam} & 59.6 & 52.8 & \textbf{59.6} & 45.8 & 54.5 \\
\midrule
\rowcolor{TableOurs}
PreWAM & \textbf{73.3} & \textbf{59.5} & 55.2 & 46.8 & \textbf{58.7} \\
\bottomrule
\end{tabularx}
\end{table}

\begin{table}[!tbp]
\centering
\caption{LIBERO-Pro2.0 success rates (\%). Base evaluates the eight unperturbed tasks in LIBERO-Pro2.0; Static and Runtime evaluate perturbations before and during execution. $\Delta=\mathrm{Success}-\mathrm{Base}$, in percentage points.}
\label{tab:pro2}
\footnotesize
\setlength{\tabcolsep}{2pt}
\begin{tabularx}{\columnwidth}{@{}l*{5}{>{\raggedleft\arraybackslash}X}@{}}
\toprule
\rowcolor{TableHeader}
Method & Base & \multicolumn{2}{c}{Static} & \multicolumn{2}{c}{Runtime} \\
\cmidrule(lr){3-4}\cmidrule(lr){5-6}
\rowcolor{TableHeader}
 & & Success & $\Delta$ & Success & $\Delta$ \\
\midrule
OpenVLA \citep{openvla} & 64.6 & 33.9 & -30.7 & 36.4 & -28.2 \\
OpenVLA-OFT \citep{openvlaoft} & 97.1 & 58.9 & -38.2 & 64.6 & -32.5 \\
$\pi_0$ \citep{pi0} & 63.8 & 38.6 & \textbf{-25.1} & 35.3 & -28.4 \\
$\pi_0$-FAST \citep{pifast} & 80.4 & 53.2 & -27.3 & 55.9 & -24.5 \\
$\pi_{0.5}$ \citep{pi05} & 93.3 & \textbf{66.7} & -26.6 & \textbf{73.8} & \textbf{-19.5} \\
UniVLA \citep{univla} & 92.1 & 52.9 & -39.2 & 61.0 & -31.1 \\
X-VLA \citep{xvla} & 95.0 & 61.2 & -33.8 & 66.0 & -29.0 \\
GR00T N1.7 \citep{gr00t17} & 90.4 & 60.0 & -30.4 & 63.9 & -26.6 \\
Fast-WAM \citep{fastwam} & \textbf{97.8} & 55.9 & -41.8 & 70.1 & -27.6 \\
\midrule
\rowcolor{TableOurs}
\textbf{PreWAM} & 92.1 & 64.0 & -28.1 & 70.1 & -22.0 \\
\bottomrule
\end{tabularx}
\end{table}

\begin{table*}[tp]
\centering
\caption{SimplerEnv Visual Matching success rates (\%). \textdagger: averages recomputed from reported task scores.}
\label{tab:simpler}
\footnotesize
\setlength{\tabcolsep}{4pt}
\begin{tabular}{@{}>{\raggedright\arraybackslash}p{0.28\textwidth}rrrrrrrrrr@{}}
\toprule
\rowcolor{TableHeader}
 & \multicolumn{5}{c}{Google Robot} & \multicolumn{5}{c}{WidowX Robot} \\
\cmidrule(lr){2-6}\cmidrule(lr){7-11}
\rowcolor{TableHeader}
Method & Pick & Move & Drawer & Place & Avg. & Spoon & Carrot & Block & Eggplant & Avg. \\
\midrule
GR00T N1 \citep{gr00t} & 0.7 & 1.9 & 2.9 & 0.0 & 1.4 & 1.4 & 0.0 & 0.0 & 13.9 & 3.8 \\
VLA-JEPA \citep{vlajepa} & 88.3 & 64.1 & 59.3 & 49.1 & 65.2 & 75.0 & \textbf{70.8} & 12.5 & 70.8 & 57.3 \\
RoboVLMs \citep{robovlms} & 77.3 & 61.7 & 43.5 & 24.1 & 51.7 & 45.8 & 20.8 & 4.2 & 79.2 & 37.5 \\
villa-X \citep{villax} & 81.7 & 55.4 & 38.4 & 4.2 & 44.9 & 48.3 & 24.2 & 19.2 & 71.7 & 40.8 \\
CogACT \citep{cogact} & 91.3 & \textbf{85.0} & 71.8 & 50.9 & 74.8 & 71.7 & 50.8 & 15.0 & 67.5 & 51.3 \\
StarVLA-OFT \citep{starvla} & \textbf{95.3} & 75.0 & 68.8 & 66.1 & 76.3\textsuperscript{\textdagger} & \textbf{90.3} & 38.5 & 29.7 & \textbf{100.0} & 64.6 \\
VideoVLA \citep{videovla} & 92.3 & 82.9 & 66.2 & 50.9 & 73.1 & 75.0 & 20.8 & \textbf{45.8} & 70.8 & 53.1 \\
\midrule
\rowcolor{TableOurs}
\textbf{PreWAM} & 91.3 & 73.3 & \textbf{75.0} & \textbf{75.9} & \textbf{78.9} & 83.3 & 58.3 & 37.5 & \textbf{100.0} & \textbf{69.8} \\
\bottomrule
\end{tabular}
\end{table*}

\begin{table}[!tbp]
\centering
\caption{RoboTwin 2.0 success rates (\%).}
\label{tab:robotwin}
\footnotesize
\setlength{\tabcolsep}{3pt}
\begin{tabularx}{\columnwidth}{@{}l*{3}{>{\raggedleft\arraybackslash}X}@{}}
\toprule
\rowcolor{TableHeader}
Method & Clean & Rand. & Avg. \\
\midrule
$\pi_0$ \citep{pi0} & 65.92 & 58.40 & 62.16 \\
$\pi_{0.5}$ \citep{pi05} & 82.74 & 76.76 & 79.75 \\
X-VLA \citep{xvla} & 72.80 & 72.84 & 72.82 \\
LingBot-VLA \citep{lingbotvla} & 88.56 & 86.68 & 87.62 \\
\midrule
Motus \citep{motus} & 88.66 & 87.02 & 87.84 \\
GigaWorld-Policy \citep{gigaworld} & 86.36 & 85.04 & 85.70 \\
LaWAM \citep{lawam} & 92.64 & 89.80 & 91.22 \\
LingBot-VA \citep{lingbotva} & \textbf{92.93} & 91.55 & 92.24 \\
\midrule
\rowcolor{TableOurs}
\textbf{PreWAM} & 92.72 & \textbf{91.98} & \textbf{92.35} \\
\bottomrule
\end{tabularx}
\end{table}

\begin{table}[!tbp]
\centering
\caption{Ablation of predictive supervision and pretraining. We report success rates (\%) on LIBERO, its Long suite, and LIBERO-Plus. WM denotes world modeling; w/o WM removes predictive supervision from both training stages. The variant without pretraining retains the full downstream objectives.}
\label{tab:ablations}
\label{tab:variants}
\small
\setlength{\tabcolsep}{3pt}
\begin{tabularx}{\columnwidth}{@{}>{\raggedright\arraybackslash}Xrrr@{}}
\toprule
\rowcolor{TableHeader}
Configuration & LIBERO & LIBERO-Long & LIBERO-Plus \\
\midrule
\rowcolor{TableOurs}
\textbf{PreWAM} & \textbf{98.1} & \textbf{97.0} & \textbf{79.1} \\
w/o TSP & 97.6 & 95.0 & 77.7 \\
w/o RWP \& TSP & 97.3 & 95.0 & 76.6 \\
\midrule
w/o WM & 96.7 & 92.6 & 72.7 \\
w/o pretraining & 95.6 & 89.8 & 65.9 \\
\bottomrule
\end{tabularx}
\end{table}

\subsection{Training and Inference}
Action learning and world prediction jointly train the shared backbone, with the available supervision determined by the data source.

\textbf{Pretraining.}
We initialize the backbone from a pretrained VLM and learn the world predictor and action expert jointly. Pretraining alternates robot demonstrations with human videos. Robot demonstrations supply both action and visual transition targets, giving the joint objective
\begin{equation}
\mathcal L_{\mathrm{robot}}=
\mathcal L_{\mathrm{act}}+\lambda_{\mathrm{wm}}\mathcal L_{\mathrm{wm}}.
\end{equation}
Human videos provide the same visual transition targets without robot-action labels and are optimized with $\mathcal L_{\mathrm{wm}}$ alone.  We set $\lambda_{\mathrm{wm}}=0.1$ and $\beta=0.5$.

\textbf{Posttraining.}
The downstream policy inherits the backbone, predictor, action expert, and learned queries from pretraining. It is then trained on demonstrations from the target manipulation setting with $\mathcal L_{\mathrm{robot}}$.  The target encoder remains frozen in both stages.

\textbf{Inference.}
Current observations and the instruction pass through the VLM. Conditioned on the embodied latent actions and available proprioception, the action expert integrates its learned velocity field from noise to an action chunk. World queries remain in the backbone input, preserving the token context used during training. The frozen target encoder and the world predictor are not executed.

\section{Experiments}\label{sec:experiments}

\subsection{Experimental Setup}
\textbf{Training settings.}
PreWAM uses Qwen3-VL-2B-Instruct \citep{qwen}, a 16-layer transformer action expert, and frozen DINOv3 ViT-L/16 targets. Pretraining alternates DROID robot demonstrations \citep{droid} with Something-Something V2 human videos \citep{ssv2} for 6,250 cycles, each comprising one robot update and one video update. The global batch size is 1,024 per update. We then posttrain separate policies on LIBERO demonstrations and on the Bridge/RT-1 mixture for SimplerEnv, both with a global batch size of 256. For LIBERO and SimplerEnv, downstream training retains both action and world objectives and runs for 36,000 updates. The representation comparisons and ablations use the same downstream batch size, update count, and optimization settings. For RoboTwin 2.0, we posttrain one policy on 50 tasks using 50 clean and 500 randomized demonstrations per task, with a global batch size of 256 for 100,000 updates. The policy predicts and executes 32-step action chunks with four flow-matching inference steps; LIBERO and SimplerEnv use seven-step chunks. The real-world setup and evaluation protocol are described in Sec.~\ref{sec:real-world}.

\textbf{Benchmarks and protocols.}
LIBERO tests four suites of language-conditioned manipulation \citep{libero}. Its robustness extensions evaluate the learned policy without adaptation: LIBERO-Plus introduces seven categories of distribution shift \citep{plus}, and LIBERO-Pro changes objects, positions, semantics, tasks, and environments \citep{pro}. LIBERO-Pro2.0 further separates static changes from interventions triggered during execution \citep{pro2}. SimplerEnv provides real-to-sim evaluation of policies trained on real robot demonstrations \citep{simpler}. We use its Visual Matching setting for Google and WidowX robots, which aligns simulated scenes with the corresponding real environments. RoboTwin 2.0 tests bimanual manipulation under clean and randomized conditions \citep{robotwin2}. We evaluate 100 episodes per task in each setting. Success rate is the common evaluation metric, with benchmark-specific aggregation: suite averages for LIBERO and Pro, category averages for Plus, official Static and Runtime aggregates for Pro2.0, separate task-category averages for the two SimplerEnv robots, and average task success under each RoboTwin 2.0 condition. Baseline results on simulation benchmarks are taken from the corresponding papers and official benchmark reports.

\subsection{Benchmark Results}
\textbf{LIBERO Suite.}
PreWAM maintains strong performance on standard manipulation tasks and generalizes to distribution shifts without additional adaptation. It achieves 98.1\% on LIBERO, close to Fast-WAM and Motus, but reaches 79.1\% on LIBERO-Plus, exceeding them by 27.6 and 34.4 percentage points (Table~\ref{tab:libero}). This contrast shows that similar performance on standard tasks can conceal substantial differences in robustness. PreWAM achieves the highest success rates in the Robot, Language, and Light categories, with strengths spanning several types of variation.

On LIBERO-Pro, PreWAM achieves the highest average success rate of 58.7\%, leading the Spatial and Object suites (Table~\ref{tab:pro}). LIBERO-Pro2.0 further evaluates static perturbations and interventions during execution. PreWAM achieves 64.0\% and 70.1\% in these settings, with drops of 28.1 and 22.0 percentage points from its unperturbed performance (Table~\ref{tab:pro2}). Its Runtime drop is the second smallest after $\pi_{0.5}$, while its Static drop is also among the smallest, showing strong retention of manipulation performance under both types of disturbance.

\textbf{SimplerEnv.}
PreWAM also performs well in real-to-sim evaluation, achieving the highest average success rates of 78.9\% on Google Robot and 69.8\% on WidowX (Table~\ref{tab:simpler}). On Google Robot, it leads on Drawer and Place, reaching 75.0\% and 75.9\%. On WidowX, it achieves 100.0\% on Eggplant while maintaining strong performance on Spoon and Block. These results extend the effectiveness of PreWAM to policies trained on real robot demonstrations and evaluated in matched simulation environments.

\textbf{RoboTwin 2.0.}
PreWAM scales to multitask bimanual manipulation with a single policy across 50 tasks. It achieves 92.72\% on Clean and 91.98\% on Randomized, averaging 92.35\% (Table~\ref{tab:robotwin}). Its Clean performance is comparable to LingBot-VA and LaWAM, while its Randomized success rate is the highest among the compared methods. The 0.74-point gap between the two settings indicates consistent performance under scene randomization in this multitask bimanual benchmark.

% Queue the analysis figure before the real-world figure.
\begin{figure}[t]
    \centering
    \includegraphics[width=\columnwidth]{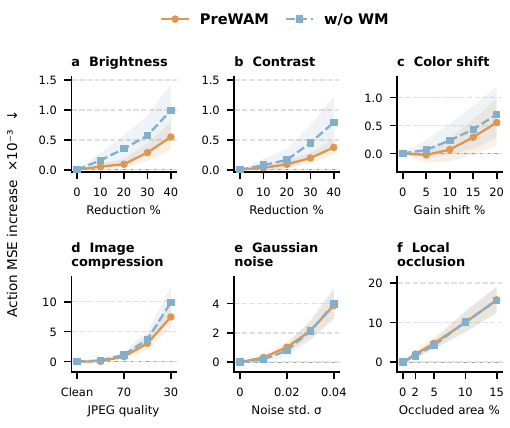}
    \caption{\textbf{Action stability under visual perturbations.} Normalized action MSE increase over clean inputs. Shading: task-stratified 95\% bootstrap confidence intervals. Vertical scales vary.}
    \label{fig:robustness}
\end{figure}

\begin{figure*}[t]
    \centering
    \includegraphics[width=\textwidth]{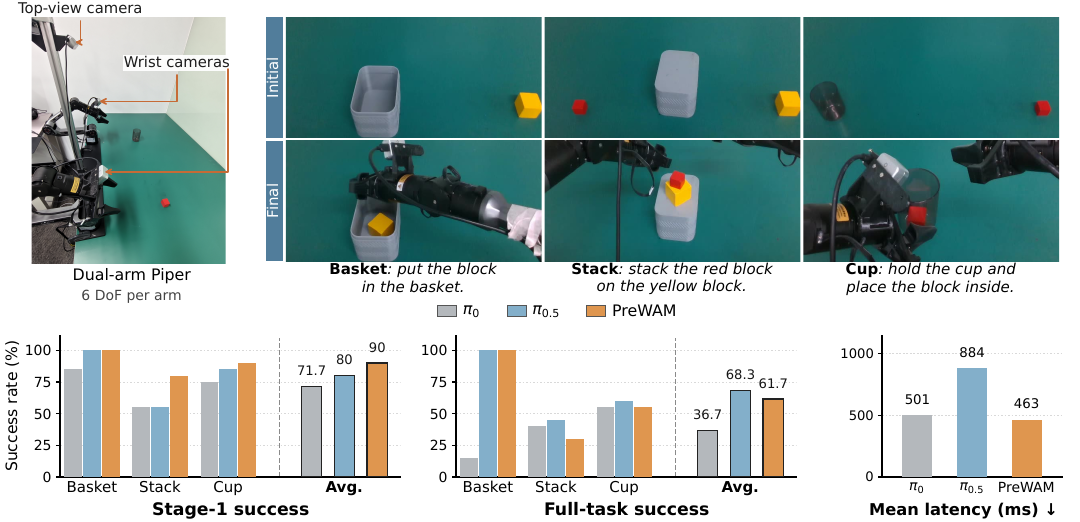}
    \caption{\textbf{Real-world evaluation on a dual-arm Piper.} Top: setup and initial/final demonstration states. Bottom: stage-1 and full-task success over 20 trials per task, and mean deployment inference latency. \textit{Avg.}: mean across three tasks.}
    \label{fig:real-world}
\end{figure*}

\subsection{Ablation Studies and Analysis}
\textbf{Effects of key designs.}
Table~\ref{tab:ablations} compares direct future-state prediction, single-horizon residual prediction, and the full multi-horizon model, alongside ablations of world modeling and pretraining. Replacing direct future-state prediction with RWP improves success on LIBERO-Plus from 76.6\% to 77.7\% and on LIBERO from 97.3\% to 97.6\%. Adding TSP brings further gains, raising LIBERO-Long success from 95.0\% to 97.0\% and LIBERO-Plus from 77.7\% to 79.1\%. Together, these results support residual prediction across multiple temporal horizons for long-horizon manipulation and generalization. Compared with action-only training, the full training recipe improves LIBERO-Long and LIBERO-Plus by 4.4 and 6.4 points, respectively, demonstrating that visual transition learning complements action supervision. With the same downstream objectives, pretraining further provides gains of 7.2 and 13.2 points on these benchmarks, highlighting the value of transferring pretrained representations to downstream policy learning.

\textbf{Action stability under visual perturbations.}
We further examine whether world modeling helps maintain action prediction accuracy under visual changes. We analyze 200 seven-step windows from LIBERO training demonstrations, covering four suites with five episodes per task and one window per episode. Both camera views are perturbed, with samples, target actions, and flow-sampling seeds fixed across models. Figure~\ref{fig:robustness} reports the increase over each policy's clean action MSE, averaged across steps and action dimensions.

The full model and the variant without world modeling have similar errors on clean inputs, but PreWAM generally exhibits smaller error increases under brightness, contrast, color-shift, and image-compression perturbations. At 40\% brightness and contrast reduction, its error increases are 44.8\% and 52.9\% lower, respectively; Gaussian noise and local occlusion yield similar responses across the two policies. These results show that training-time world modeling can improve robustness to some visual changes, with these benefits retained when inference runs only the backbone and action expert.

\subsection{Real-World Evaluation}
\label{sec:real-world}

\textbf{Setup and protocol.}
We evaluate PreWAM on a dual-arm Piper platform with two 6-DoF arms, parallel grippers, a top-view RGB camera, and two wrist cameras. Three tasks cover single-arm pick-and-place, precise bimanual placement, and coordinated bimanual manipulation (Fig.~\ref{fig:real-world}). \textit{Basket} requires one arm to place a block in a basket; stage-1 success requires grasping and lifting the block off the table. \textit{Stack} requires the two arms to position a yellow block and align and stack a red block on top; stage-1 success requires positioning the yellow block. \textit{Cup} requires the left arm to stabilize a cup while the right arm places a red block inside; stage-1 success requires stabilizing the cup with the left arm while grasping and lifting the block with the right.

We collect 50 demonstrations per task at 20\,Hz and fine-tune PreWAM, $\pi_0$~\citep{pi0}, and $\pi_{0.5}$~\citep{pi05} separately for each task. All three models undergo 4,000 task-specific fine-tuning updates. Each model is evaluated over 20 full-task trials per task, recording both stage-1 and full-task success with all 20 trials as the denominator. All policies use three current RGB views and 10 sampling steps. Actions are executed at 20 Hz with a nominal replanning interval of 16 control steps. The next chunk is prefetched eight steps in advance while the current chunk continues to execute.

\textbf{Results.}
As shown in Fig.~\ref{fig:real-world}, PreWAM achieves 90.0\% mean stage-1 success, compared with 71.7\% for $\pi_0$ and 80.0\% for $\pi_{0.5}$. The largest stage-1 gain occurs on \textit{Stack}, where PreWAM reaches 80\%, compared with 55\% for both baselines. PreWAM completes \textit{Basket} in all 20 trials and achieves 30\% and 55\% full-task success on \textit{Stack} and \textit{Cup}, respectively. These results show that PreWAM reliably completes preparatory operations such as grasping and positioning, while subsequent stacking and placement remain challenges for completing the bimanual tasks. Mean full-task success is 61.7\%, compared with 36.7\% for $\pi_0$ and 68.3\% for $\pi_{0.5}$.

\textbf{Inference efficiency.}
In real-world deployment, mean policy-call latency on an NVIDIA RTX 4090 is 463\,ms for PreWAM, 501\,ms for $\pi_0$, and 884\,ms for $\pi_{0.5}$.

\section{Conclusion}\label{sec:conclusion}
\enlargethispage{\baselineskip}
Our controlled study links prediction spaces to policy generalization: spatial semantic supervision provides stronger overall robustness and more accessible motion information under camera shifts and noise. PreWAM combines source-relative residual prediction with temporal-scale conditioning to shape shared policy representations during pretraining and posttraining, while retaining direct action generation at deployment. Evaluations on LIBERO benchmarks, SimplerEnv, and RoboTwin 2.0 support this design across tasks and embodiments. Real-world evaluations on three Piper tasks further assess single-arm and bimanual manipulation through stage-level success, full-task completion, and inference latency.

\section*{LLM Usage Statement}
OpenAI ChatGPT and Codex assisted with drafting and revising the manuscript, preparing code and visualizations, and generating figure illustrations. The authors are responsible for the scientific content and conclusions of this work.

{\small
\bibliographystyle{ieeenat_fullname}
\bibliography{references}

\begin{thebibliography}{39}
\providecommand{\natexlab}[1]{#1}
\providecommand{\url}[1]{\texttt{#1}}
\expandafter\ifx\csname urlstyle\endcsname\relax
  \providecommand{\doi}[1]{doi: #1}\else
  \providecommand{\doi}{doi: \begingroup \urlstyle{rm}\Url}\fi

\bibitem[Assran et~al.(2025)Assran, Bardes, Fan, Garrido, Howes, Komeili,
  Muckley, Rizvi, Roberts, Sinha, Zholus, Arnaud, Gejji, Martin, Hogan, Dugas,
  Bojanowski, Khalidov, Labatut, Massa, Szafraniec, Krishnakumar, Li, Ma,
  Chandar, Meier, LeCun, Rabbat, and Ballas]{vjepa2}
Mido Assran, Adrien Bardes, David Fan, Quentin Garrido, Russell Howes, Mojtaba
  Komeili, Matthew Muckley, Ammar Rizvi, Claire Roberts, Koustuv Sinha, Artem
  Zholus, Sergio Arnaud, Abha Gejji, Ada Martin, Francois~Robert Hogan, Daniel
  Dugas, Piotr Bojanowski, Vasil Khalidov, Patrick Labatut, Francisco Massa,
  Marc Szafraniec, Kapil Krishnakumar, Yong Li, Xiaodong Ma, Sarath Chandar,
  Franziska Meier, Yann LeCun, Michael Rabbat, and Nicolas Ballas.
\newblock {V-JEPA 2: Self-Supervised Video Models Enable Understanding,
  Prediction and Planning}.
\newblock \emph{arXiv preprint arXiv:2506.09985}, 2025.

\bibitem[Bai et~al.(2025)Bai, Cai, Chen, Chen, Chen, Cheng, Deng, Ding, Gao,
  Ge, Ge, Guo, Huang, Huang, Huang, Hui, Jiang, Li, Li, Li, Li, Lin, Lin, Liu,
  Liu, Liu, Liu, Liu, Liu, Lu, Luo, Lv, Men, Meng, Ren, Ren, Song, Sun, Tang,
  Tu, Wan, Wang, Wang, Wang, Wang, Xie, Xu, Xu, Xu, Yang, Yang, Yang, Yang, Yu,
  Zhang, Zhang, Zhang, Zheng, Zhong, Zhou, Zhou, Zhou, Zhu, and Zhu]{qwen}
Shuai Bai, Yuxuan Cai, Ruizhe Chen, Keqin Chen, Xionghui Chen, Zesen Cheng,
  Lianghao Deng, Wei Ding, Chang Gao, Chunjiang Ge, Wenbin Ge, Zhifang Guo,
  Qidong Huang, Jie Huang, Fei Huang, Binyuan Hui, Shutong Jiang, Zhaohai Li,
  Mingsheng Li, Mei Li, Kaixin Li, Zicheng Lin, Junyang Lin, Xuejing Liu,
  Jiawei Liu, Chenglong Liu, Yang Liu, Dayiheng Liu, Shixuan Liu, Dunjie Lu,
  Ruilin Luo, Chenxu Lv, Rui Men, Lingchen Meng, Xuancheng Ren, Xingzhang Ren,
  Sibo Song, Yuchong Sun, Jun Tang, Jianhong Tu, Jianqiang Wan, Peng Wang,
  Pengfei Wang, Qiuyue Wang, Yuxuan Wang, Tianbao Xie, Yiheng Xu, Haiyang Xu,
  Jin Xu, Zhibo Yang, Mingkun Yang, Jianxin Yang, An Yang, Bowen Yu, Fei Zhang,
  Hang Zhang, Xi Zhang, Bo Zheng, Humen Zhong, Jingren Zhou, Fan Zhou, Jing
  Zhou, Yuanzhi Zhu, and Ke Zhu.
\newblock {Qwen3-VL Technical Report}.
\newblock \emph{arXiv preprint arXiv:2511.21631}, 2025.

\bibitem[Baldassarre et~al.(2025)Baldassarre, Szafraniec, Terver, Khalidov,
  Massa, LeCun, Labatut, Seitzer, and Bojanowski]{dinoworld}
Federico Baldassarre, Marc Szafraniec, Basile Terver, Vasil Khalidov, Francisco
  Massa, Yann LeCun, Patrick Labatut, Maximilian Seitzer, and Piotr Bojanowski.
\newblock {Back to the Features: DINO as a Foundation for Video World Models}.
\newblock \emph{arXiv preprint arXiv:2507.19468}, 2025.

\bibitem[Bi et~al.(2025)Bi, Tan, Xie, Wang, Huang, Liu, Zhao, Feng, Xiang,
  Rong, Zhao, Liu, Su, Ma, Su, and Zhu]{motus}
Hongzhe Bi, Hengkai Tan, Shenghao Xie, Zeyuan Wang, Shuhe Huang, Haitian Liu,
  Ruowen Zhao, Yao Feng, Chendong Xiang, Yinze Rong, Hongyan Zhao, Hanyu Liu,
  Zhizhong Su, Lei Ma, Hang Su, and Jun Zhu.
\newblock {Motus: A Unified Latent Action World Model}.
\newblock \emph{arXiv preprint arXiv:2512.13030}, 2025.

\bibitem[Black et~al.(2025)Black, Brown, Driess, Esmail, Equi, Finn, Fusai,
  Groom, Hausman, Ichter, Jakubczak, Jones, Ke, Levine, Li-Bell, Mothukuri,
  Nair, Pertsch, Shi, Smith, Tanner, Vuong, Walling, Wang, and Zhilinsky]{pi0}
Kevin Black, Noah Brown, Danny Driess, Adnan Esmail, Michael~Robert Equi,
  Chelsea Finn, Niccolo Fusai, Lachy Groom, Karol Hausman, Brian Ichter, Szymon
  Jakubczak, Tim Jones, Liyiming Ke, Sergey Levine, Adrian Li-Bell, Mohith
  Mothukuri, Suraj Nair, Karl Pertsch, Lucy~Xiaoyang Shi, Laura Smith, James
  Tanner, Quan Vuong, Anna Walling, Haohuan Wang, and Ury Zhilinsky.
\newblock {$\pi_0$: A Vision-Language-Action Flow Model for General Robot
  Control}.
\newblock In \emph{Proceedings of Robotics: Science and Systems}, 2025.

\bibitem[Bu et~al.(2025)Bu, Yang, Cai, Gao, Ren, Yao, Luo, and Li]{univla}
Qingwen Bu, Yanting Yang, Jisong Cai, Shenyuan Gao, Guanghui Ren, Maoqing Yao,
  Ping Luo, and Hongyang Li.
\newblock {Learning to Act Anywhere with Task-centric Latent Actions}.
\newblock In \emph{Proceedings of Robotics: Science and Systems}, 2025.

\bibitem[Chen et~al.(2026{\natexlab{a}})Chen, Wang, Chen, Chen, Gao, Tang, Li,
  Liu, Yao, Li, Xu, and Yu]{lawam}
Jialei Chen, Kai Wang, Kang Chen, Shuaihang Chen, Feng Gao, Wenhao Tang,
  Zhiyuan Li, Weilin Liu, Zhuyu Yao, Boxun Li, Yuanbo Xu, and Chao Yu.
\newblock {LaWAM: Latent World Action Models for Efficient Dynamics-Aware Robot
  Policies}.
\newblock \emph{arXiv preprint arXiv:2606.15768}, 2026{\natexlab{a}}.

\bibitem[Chen et~al.(2026{\natexlab{b}})Chen, Chen, Chen, Cai, Liu, Li, Liang,
  Lin, Ge, Gu, Deng, Guo, Nian, Xie, Chen, Su, Xu, Liu, Hu, Gao, Wang, Liang,
  Qin, Yang, Luo, and Mu]{robotwin2}
Tianxing Chen, Zanxin Chen, Baijun Chen, Zijian Cai, Yibin Liu, Zixuan Li,
  Qiwei Liang, Xianliang Lin, Yiheng Ge, Zhenyu Gu, Weiliang Deng, Yubin Guo,
  Tian Nian, Xuanbing Xie, Qiangyu Chen, Kailun Su, Tianling Xu, Guodong Liu,
  Mengkang Hu, Huan-ang Gao, Kaixuan Wang, Zhixuan Liang, Yusen Qin, Xiaokang
  Yang, Ping Luo, and Yao Mu.
\newblock {RoboTwin 2.0}: A scalable data generator and benchmark with strong
  domain randomization for robust bimanual robotic manipulation.
\newblock In \emph{Proceedings of the 43rd International Conference on Machine
  Learning}, pages 13673--13699. PMLR, 2026{\natexlab{b}}.

\bibitem[Chen et~al.(2025)Chen, Wei, Zhang, Zhang, Wang, Guo, Yang, Wang, Xiao,
  Zhao, Chen, and Bian]{villax}
Xiaoyu Chen, Hangxing Wei, Pushi Zhang, Chuheng Zhang, Kaixin Wang, Yanjiang
  Guo, Rushuai Yang, Yucen Wang, Xinquan Xiao, Li Zhao, Jianyu Chen, and Jiang
  Bian.
\newblock {villa-X: Enhancing Latent Action Modeling in Vision-Language-Action
  Models}.
\newblock \emph{arXiv preprint arXiv:2507.23682}, 2025.

\bibitem[Fang et~al.(2026)Fang, Liu, Lai, Wang, Jia, Sheng, Wan, Xu, and
  Gao]{bridgewam}
Jiaxin Fang, Ruitong Liu, Zeyu Lai, Xingyu Wang, Mingyuan Jia, Pingyue Sheng,
  Weikang Wan, Mengdi Xu, and Yuxiang Gao.
\newblock {BridgeWAM: Bridging Video and Action Experts via Latent Bridge
  Queries}.
\newblock \emph{arXiv preprint arXiv:2610.14746}, 2026.

\bibitem[Fei et~al.(2025)Fei, Wang, Shi, Dai, Cai, Qian, Ji, He, Zhang, Fei,
  Fu, Gong, and Qiu]{plus}
Senyu Fei, Siyin Wang, Junhao Shi, Zihao Dai, Jikun Cai, Pengfang Qian, Li Ji,
  Xinzhe He, Shiduo Zhang, Zhaoye Fei, Jinlan Fu, Jingjing Gong, and Xipeng
  Qiu.
\newblock {LIBERO-Plus: In-depth Robustness Analysis of Vision-Language-Action
  Models}.
\newblock \emph{arXiv preprint arXiv:2510.13626}, 2025.

\bibitem[Goyal et~al.(2017)Goyal, Kahou, Michalski, Materzynska, Westphal, Kim,
  Haenel, Fruend, Yianilos, Mueller-Freitag, Hoppe, Thurau, Bax, and
  Memisevic]{ssv2}
Raghav Goyal, Samira~Ebrahimi Kahou, Vincent Michalski, Joanna Materzynska,
  Susanne Westphal, Heuna Kim, Valentin Haenel, Ingo Fruend, Peter Yianilos,
  Moritz Mueller-Freitag, Florian Hoppe, Christian Thurau, Ingo Bax, and Roland
  Memisevic.
\newblock {The ``Something Something Video Database for Learning and Evaluating
  Visual Common Sense}.
\newblock In \emph{Proceedings of the IEEE International Conference on Computer
  Vision}, 2017.

\bibitem[Khazatsky et~al.(2024)Khazatsky, Pertsch, Nair, Balakrishna, Dasari,
  Karamcheti, Nasiriany, Srirama, Chen, Ellis, Fagan, Hejna, Itkina, Lepert,
  Ma, Miller, Wu, Belkhale, Dass, Ha, Jain, Lee, Lee, Memmel, Park,
  Radosavovic, Wang, Zhan, Black, Chi, Hatch, Lin, Lu, Mercat, Rehman, Sanketi,
  Sharma, Simpson, Vuong, Walke, Wulfe, Xiao, Yang, Yavary, Zhao, Agia, Baijal,
  Castro, Chen, Chen, Chung, Drake, Foster, Gao, Guizilini, Herrera, Heo, Hsu,
  Hu, Irshad, Jackson, Le, Li, Lin, Lin, Ma, Maddukuri, Mirchandani, Morton,
  Nguyen, O'Neill, Scalise, Seale, Son, Tian, Tran, Wang, Wu, Xie, Yang, Yin,
  Zhang, Bastani, Berseth, Bohg, Goldberg, Gupta, Gupta, Jayaraman, Lim, Malik,
  Mart{\'i}n-Mart{\'i}n, Ramamoorthy, Sadigh, Song, Wu, Yip, Zhu, Kollar,
  Levine, and Finn]{droid}
Alexander Khazatsky, Karl Pertsch, Suraj Nair, Ashwin Balakrishna, Sudeep
  Dasari, Siddharth Karamcheti, Soroush Nasiriany, Mohan~Kumar Srirama,
  Lawrence~Yunliang Chen, Kirsty Ellis, Peter~David Fagan, Joey Hejna, Masha
  Itkina, Marion Lepert, Yecheng~Jason Ma, Patrick~Tree Miller, Jimmy Wu,
  Suneel Belkhale, Shivin Dass, Huy Ha, Arhan Jain, Abraham Lee, Youngwoon Lee,
  Marius Memmel, Sungjae Park, Ilija Radosavovic, Kaiyuan Wang, Albert Zhan,
  Kevin Black, Cheng Chi, Kyle~Beltran Hatch, Shan Lin, Jingpei Lu, Jean
  Mercat, Abdul Rehman, Pannag~R Sanketi, Archit Sharma, Cody Simpson, Quan
  Vuong, Homer~Rich Walke, Blake Wulfe, Ted Xiao, Jonathan~Heewon Yang, Arefeh
  Yavary, Tony~Z. Zhao, Christopher Agia, Rohan Baijal, Mateo~Guaman Castro,
  Daphne Chen, Qiuyu Chen, Trinity Chung, Jaimyn Drake, Ethan~Paul Foster,
  Jensen Gao, Vitor Guizilini, David~Antonio Herrera, Minho Heo, Kyle Hsu,
  Jiaheng Hu, Muhammad~Zubair Irshad, Donovon Jackson, Charlotte Le, Yunshuang
  Li, Kevin Lin, Roy Lin, Zehan Ma, Abhiram Maddukuri, Suvir Mirchandani,
  Daniel Morton, Tony Nguyen, Abigail O'Neill, Rosario Scalise, Derick Seale,
  Victor Son, Stephen Tian, Emi Tran, Andrew~E. Wang, Yilin Wu, Annie Xie,
  Jingyun Yang, Patrick Yin, Yunchu Zhang, Osbert Bastani, Glen Berseth,
  Jeannette Bohg, Ken Goldberg, Abhinav Gupta, Abhishek Gupta, Dinesh
  Jayaraman, Joseph~J Lim, Jitendra Malik, Roberto Mart{\'i}n-Mart{\'i}n,
  Subramanian Ramamoorthy, Dorsa Sadigh, Shuran Song, Jiajun Wu, Michael~C.
  Yip, Yuke Zhu, Thomas Kollar, Sergey Levine, and Chelsea Finn.
\newblock {DROID: A Large-Scale In-The-Wild Robot Manipulation Dataset}.
\newblock \emph{arXiv preprint arXiv:2403.12945}, 2024.

\bibitem[Kim et~al.(2025{\natexlab{a}})Kim, Finn, and Liang]{openvlaoft}
Moo~Jin Kim, Chelsea Finn, and Percy Liang.
\newblock {Fine-Tuning Vision-Language-Action Models: Optimizing Speed and
  Success}.
\newblock In \emph{Proceedings of Robotics: Science and Systems},
  2025{\natexlab{a}}.

\bibitem[Kim et~al.(2025{\natexlab{b}})Kim, Pertsch, Karamcheti, Xiao,
  Balakrishna, Nair, Rafailov, Foster, Sanketi, Vuong, Kollar, Burchfiel,
  Tedrake, Sadigh, Levine, Liang, and Finn]{openvla}
Moo~Jin Kim, Karl Pertsch, Siddharth Karamcheti, Ted Xiao, Ashwin Balakrishna,
  Suraj Nair, Rafael Rafailov, Ethan~P Foster, Pannag~R Sanketi, Quan Vuong,
  Thomas Kollar, Benjamin Burchfiel, Russ Tedrake, Dorsa Sadigh, Sergey Levine,
  Percy Liang, and Chelsea Finn.
\newblock {OpenVLA: An Open-Source Vision-Language-Action Model}.
\newblock In \emph{Proceedings of The 8th Conference on Robot Learning}, pages
  2679--2713. PMLR, 2025{\natexlab{b}}.

\bibitem[Li et~al.(2026)Li, Zhang, Luo, Yang, Wang, Han, Yu, Gao, Xue, Zhu,
  Shen, and Xu]{lingbotva}
Lin Li, Qihang Zhang, Yiming Luo, Shuai Yang, Ruilin Wang, Fei Han, Mingrui Yu,
  Zelin Gao, Nan Xue, Xing Zhu, Yujun Shen, and Yinghao Xu.
\newblock {Causal World Modeling for Robot Control}.
\newblock \emph{arXiv preprint arXiv:2601.21998}, 2026.

\bibitem[Li et~al.(2024{\natexlab{a}})Li, Liang, Wang, Luo, Chen, Liao, Wei,
  Deng, Xu, Zhang, Wang, Liu, Fu, Bao, Chen, Shi, Yang, and Guo]{cogact}
Qixiu Li, Yaobo Liang, Zeyu Wang, Lin Luo, Xi Chen, Mozheng Liao, Fangyun Wei,
  Yu Deng, Sicheng Xu, Yizhong Zhang, Xiaofan Wang, Bei Liu, Jianlong Fu,
  Jianmin Bao, Dong Chen, Yuanchun Shi, Jiaolong Yang, and Baining Guo.
\newblock {CogACT}: A foundational vision-language-action model for synergizing
  cognition and action in robotic manipulation.
\newblock \emph{arXiv preprint arXiv:2411.19650}, 2024{\natexlab{a}}.

\bibitem[Li et~al.(2024{\natexlab{b}})Li, Hsu, Gu, Pertsch, Mees, Walke, Fu,
  Lunawat, Sieh, Kirmani, Levine, Wu, Finn, Su, Vuong, and Xiao]{simpler}
Xuanlin Li, Kyle Hsu, Jiayuan Gu, Karl Pertsch, Oier Mees, Homer~Rich Walke,
  Chuyuan Fu, Ishikaa Lunawat, Isabel Sieh, Sean Kirmani, Sergey Levine, Jiajun
  Wu, Chelsea Finn, Hao Su, Quan Vuong, and Ted Xiao.
\newblock {Evaluating Real-World Robot Manipulation Policies in Simulation}.
\newblock \emph{arXiv preprint arXiv:2405.05941}, 2024{\natexlab{b}}.

\bibitem[Li et~al.(2024{\natexlab{c}})Li, Li, Qian, Liu, Wang, Liu, Kang, Ma,
  Wang, Guo, Kong, Zhang, and Liu]{robovlms}
Xinghang Li, Peiyan Li, Long Qian, Minghuan Liu, Dong Wang, Jirong Liu, Bingyi
  Kang, Xiao Ma, Xinlong Wang, Di Guo, Tao Kong, Hanbo Zhang, and Huaping Liu.
\newblock {What Matters in Building Vision-Language-Action Models for
  Generalist Robots}.
\newblock \emph{arXiv preprint arXiv:2412.14058}, 2024{\natexlab{c}}.

\bibitem[Lin et~al.(2026)Lin, He, Bao, Zhao, Li, Wang, Wang, Chi, and
  Zhang]{jepawam}
Yihan Lin, Jiawei He, Shifeng Bao, Chen Zhao, Yang Li, Xiaobo Wang, Yan Wang,
  Cheng Chi, and Jing Zhang.
\newblock {JEPA-WAM: Learning Vision-Language-Action Policies with
  Joint-Embedding World Modeling}.
\newblock \emph{arXiv preprint arXiv:2608.09381}, 2026.

\bibitem[Liu et~al.(2023)Liu, Zhu, Gao, Feng, Liu, Zhu, and Stone]{libero}
Bo Liu, Yifeng Zhu, Chongkai Gao, Yihao Feng, Qiang Liu, Yuke Zhu, and Peter
  Stone.
\newblock {LIBERO: Benchmarking Knowledge Transfer for Lifelong Robot
  Learning}.
\newblock In \emph{Advances in Neural Information Processing Systems}, 2023.

\bibitem[Llontop and Vadrevu(2026)]{gr00t17}
Edith Llontop and Kalyan Vadrevu.
\newblock {NVIDIA Isaac GR00T N1.7: Open Reasoning VLA Model for Humanoid
  Robots}.
\newblock Hugging Face Blog.
  \url{https://huggingface.co/blog/nvidia/gr00t-n1-7}, 2026.

\bibitem[{NVIDIA} et~al.(2025){NVIDIA}, Bjorck, Casta{\~n}eda, Cherniadev, Da,
  Ding, Fan, Fang, Fox, Hu, Huang, Jang, Jiang, Kautz, Kundalia, Lao, Li, Lin,
  Lin, Liu, Llontop, Magne, Mandlekar, Narayan, Nasiriany, Reed, Tan, Wang,
  Wang, Wang, Wang, Xiang, Xie, Xu, Xu, Ye, Yu, Zhang, Zhang, Zhao, Zheng, and
  Zhu]{gr00t}
{NVIDIA}, Johan Bjorck, Fernando Casta{\~n}eda, Nikita Cherniadev, Xingye Da,
  Runyu Ding, Linxi~"Jim" Fan, Yu Fang, Dieter Fox, Fengyuan Hu, Spencer Huang,
  Joel Jang, Zhenyu Jiang, Jan Kautz, Kaushil Kundalia, Lawrence Lao, Zhiqi Li,
  Zongyu Lin, Kevin Lin, Guilin Liu, Edith Llontop, Loic Magne, Ajay Mandlekar,
  Avnish Narayan, Soroush Nasiriany, Scott Reed, You~Liang Tan, Guanzhi Wang,
  Zu Wang, Jing Wang, Qi Wang, Jiannan Xiang, Yuqi Xie, Yinzhen Xu, Zhenjia Xu,
  Seonghyeon Ye, Zhiding Yu, Ao Zhang, Hao Zhang, Yizhou Zhao, Ruijie Zheng,
  and Yuke Zhu.
\newblock {GR00T N1: An Open Foundation Model for Generalist Humanoid Robots}.
\newblock \emph{arXiv preprint arXiv:2503.14734}, 2025.

\bibitem[Pertsch et~al.(2025)Pertsch, Stachowicz, Ichter, Driess, Nair, Vuong,
  Mees, Finn, and Levine]{pifast}
Karl Pertsch, Kyle Stachowicz, Brian Ichter, Danny Driess, Suraj Nair, Quan
  Vuong, Oier Mees, Chelsea Finn, and Sergey Levine.
\newblock {FAST: Efficient Action Tokenization for Vision-Language-Action
  Models}.
\newblock \emph{arXiv preprint arXiv:2501.09747}, 2025.

\bibitem[{Physical Intelligence} et~al.(2025){Physical Intelligence}, Black,
  Brown, Darpinian, Dhabalia, Driess, Esmail, Equi, Finn, Fusai, Galliker,
  Ghosh, Groom, Hausman, Ichter, Jakubczak, Jones, Ke, LeBlanc, Levine,
  Li-Bell, Mothukuri, Nair, Pertsch, Ren, Shi, Smith, Springenberg, Stachowicz,
  Tanner, Vuong, Walke, Walling, Wang, Yu, and Zhilinsky]{pi05}
{Physical Intelligence}, Kevin Black, Noah Brown, James Darpinian, Karan
  Dhabalia, Danny Driess, Adnan Esmail, Michael Equi, Chelsea Finn, Niccolo
  Fusai, Manuel~Y. Galliker, Dibya Ghosh, Lachy Groom, Karol Hausman, Brian
  Ichter, Szymon Jakubczak, Tim Jones, Liyiming Ke, Devin LeBlanc, Sergey
  Levine, Adrian Li-Bell, Mohith Mothukuri, Suraj Nair, Karl Pertsch, Allen~Z.
  Ren, Lucy~Xiaoyang Shi, Laura Smith, Jost~Tobias Springenberg, Kyle
  Stachowicz, James Tanner, Quan Vuong, Homer Walke, Anna Walling, Haohuan
  Wang, Lili Yu, and Ury Zhilinsky.
\newblock {$\pi_{0.5}$: A Vision-Language-Action Model with Open-World
  Generalization}.
\newblock \emph{arXiv preprint arXiv:2504.16054}, 2025.

\bibitem[Shen et~al.(2025)Shen, Wei, Du, Liang, Lu, Yang, Zheng, and
  Guo]{videovla}
Yichao Shen, Fangyun Wei, Zhiying Du, Yaobo Liang, Yan Lu, Jiaolong Yang,
  Nanning Zheng, and Baining Guo.
\newblock {VideoVLA}: Video generators can be generalizable robot manipulators.
\newblock In \emph{Advances in Neural Information Processing Systems}, 2025.

\bibitem[Sim{\'e}oni et~al.(2025)Sim{\'e}oni, Vo, Seitzer, Baldassarre, Oquab,
  Jose, Khalidov, Szafraniec, Yi, Ramamonjisoa, Massa, Haziza, Wehrstedt, Wang,
  Darcet, Moutakanni, Sentana, Roberts, Vedaldi, Tolan, Brandt, Couprie,
  Mairal, J{\'e}gou, Labatut, and Bojanowski]{dinov3}
Oriane Sim{\'e}oni, Huy~V. Vo, Maximilian Seitzer, Federico Baldassarre, Maxime
  Oquab, Cijo Jose, Vasil Khalidov, Marc Szafraniec, Seungeun Yi, Micha{\"e}l
  Ramamonjisoa, Francisco Massa, Daniel Haziza, Luca Wehrstedt, Jianyuan Wang,
  Timoth{\'e}e Darcet, Th{\'e}o Moutakanni, Leonel Sentana, Claire Roberts,
  Andrea Vedaldi, Jamie Tolan, John Brandt, Camille Couprie, Julien Mairal,
  Herv{\'e} J{\'e}gou, Patrick Labatut, and Piotr Bojanowski.
\newblock {DINOv3}.
\newblock \emph{arXiv preprint arXiv:2508.10104}, 2025.

\bibitem[{StarVLA Community}(2026)]{starvla}
{StarVLA Community}.
\newblock {StarVLA}: A lego-like codebase for vision-language-action model
  developing.
\newblock \emph{arXiv preprint arXiv:2604.05014}, 2026.

\bibitem[Sun et~al.(2026)Sun, Zhang, Qi, Ren, Liu, Zhu, Sun, Jin, and
  Chen]{vlajepa}
Jingwen Sun, Wenyao Zhang, Zekun Qi, Shaojie Ren, Zezhi Liu, Hanxin Zhu,
  Guangzhong Sun, Xin Jin, and Zhibo Chen.
\newblock {VLA-JEPA: Enhancing Vision-Language-Action Model with Latent World
  Model}.
\newblock \emph{arXiv preprint arXiv:2602.10098}, 2026.

\bibitem[{Team Wan} et~al.(2025){Team Wan}, Wang, Ai, Wen, Mao, Xie, Chen, Yu,
  Zhao, Yang, Zeng, Wang, Zhang, Zhou, Wang, Chen, Zhu, Zhao, Yan, Huang, Feng,
  Zhang, Li, Wu, Chu, Feng, Zhang, Sun, Fang, Wang, Gui, Weng, Shen, Lin, Wang,
  Wang, Zhou, Wang, Shen, Yu, Shi, Huang, Xu, Kou, Lv, Li, Liu, Wang, Zhang,
  Huang, Li, Wu, Liu, Pan, Zheng, Hong, Shi, Feng, Jiang, Han, Wu, and
  Liu]{wan}
{Team Wan}, Ang Wang, Baole Ai, Bin Wen, Chaojie Mao, Chen-Wei Xie, Di Chen,
  Feiwu Yu, Haiming Zhao, Jianxiao Yang, Jianyuan Zeng, Jiayu Wang, Jingfeng
  Zhang, Jingren Zhou, Jinkai Wang, Jixuan Chen, Kai Zhu, Kang Zhao, Keyu Yan,
  Lianghua Huang, Mengyang Feng, Ningyi Zhang, Pandeng Li, Pingyu Wu, Ruihang
  Chu, Ruili Feng, Shiwei Zhang, Siyang Sun, Tao Fang, Tianxing Wang, Tianyi
  Gui, Tingyu Weng, Tong Shen, Wei Lin, Wei Wang, Wei Wang, Wenmeng Zhou, Wente
  Wang, Wenting Shen, Wenyuan Yu, Xianzhong Shi, Xiaoming Huang, Xin Xu, Yan
  Kou, Yangyu Lv, Yifei Li, Yijing Liu, Yiming Wang, Yingya Zhang, Yitong
  Huang, Yong Li, You Wu, Yu Liu, Yulin Pan, Yun Zheng, Yuntao Hong, Yupeng
  Shi, Yutong Feng, Zeyinzi Jiang, Zhen Han, Zhi-Fan Wu, and Ziyu Liu.
\newblock {Wan: Open and Advanced Large-Scale Video Generative Models}.
\newblock \emph{arXiv preprint arXiv:2503.20314}, 2025.

\bibitem[Wu et~al.(2026)Wu, Lu, Wang, Yang, Liu, Wang, Zhu, Sun, Wang, Ma, Ren,
  Zhang, Yu, Zhao, Zhou, Qiu, Xiong, Wang, Wang, Cheng, Li, Huang, Zhu, Shen,
  and Zheng]{lingbotvla}
Wei Wu, Fan Lu, Yunnan Wang, Shuai Yang, Shi Liu, Fangjing Wang, Qian Zhu, He
  Sun, Yong Wang, Shuailei Ma, Yiyu Ren, Kejia Zhang, Hui Yu, Jingmei Zhao,
  Shuai Zhou, Zhenqi Qiu, Houlong Xiong, Ziyu Wang, Zechen Wang, Ran Cheng,
  Yong-Lu Li, Yongtao Huang, Xing Zhu, Yujun Shen, and Kecheng Zheng.
\newblock {A Pragmatic VLA Foundation Model}.
\newblock \emph{arXiv preprint arXiv:2601.18692}, 2026.

\bibitem[Ye et~al.(2026)Ye, Wang, Ni, Huang, Zhao, Li, Li, Li, Lv, Liu, Cao,
  Li, Deng, Mei, Wang, Chen, Zhou, Wang, Chang, Li, Zhou, Ye, Liu, and
  Zhu]{gigaworld}
Angen Ye, Boyuan Wang, Chaojun Ni, Guan Huang, Guosheng Zhao, Hao Li, Hengtao
  Li, Jie Li, Jindi Lv, Jingyu Liu, Min Cao, Peng Li, Qiuping Deng, Wenjun Mei,
  Xiaofeng Wang, Xinze Chen, Xinyu Zhou, Yang Wang, Yifan Chang, Yifan Li,
  Yukun Zhou, Yun Ye, Zhichao Liu, and Zheng Zhu.
\newblock {GigaWorld-Policy: An Efficient Action-Centered World--Action Model}.
\newblock \emph{arXiv preprint arXiv:2603.17240}, 2026.

\bibitem[Yuan et~al.(2026)Yuan, Dong, Liu, and Zhao]{fastwam}
Tianyuan Yuan, Zibin Dong, Yicheng Liu, and Hang Zhao.
\newblock {Fast-WAM: Do World Action Models Need Test-time Future Imagination?}
\newblock \emph{arXiv preprint arXiv:2603.16666}, 2026.

\bibitem[Zhang et~al.(2026)Zhang, Zhang, Qi, Zhang, Lin, Zhang, Mu, Yang, Zeng,
  and Jin]{imagewam}
Yuyang Zhang, Wenyao Zhang, Zekun Qi, He Zhang, Haitao Lin, Jingbo Zhang, Yao
  Mu, Xiaokang Yang, Wenjun Zeng, and Xin Jin.
\newblock {ImageWAM: Do World Action Models Really Need Video Generation, or
  Just Image Editing?}
\newblock \emph{arXiv preprint arXiv:2606.19531}, 2026.

\bibitem[Zhao et~al.(2026)Zhao, Jiang, Shi, Liu, Huang, Su, Sui, and
  Wang]{fasterwam}
Weiheng Zhao, Haoyi Jiang, Xin Shi, Liu Liu, Fan Huang, Zhizhong Su, Wei Sui,
  and Xinggang Wang.
\newblock {Faster-WAM: Efficient Inference-Time Future Conditioning for Robust
  World Action Models}.
\newblock \emph{arXiv preprint arXiv:2608.04404}, 2026.

\bibitem[Zheng et~al.(2025)Zheng, Li, Wang, Liu, Kang, Feng, Zheng, Zou, Chen,
  Zeng, Zhang, Pang, Liu, Wang, and Zhan]{xvla}
Jinliang Zheng, Jianxiong Li, Zhihao Wang, Dongxiu Liu, Xirui Kang, Yuchun
  Feng, Yinan Zheng, Jiayin Zou, Yilun Chen, Jia Zeng, Ya-Qin Zhang, Jiangmiao
  Pang, Jingjing Liu, Tai Wang, and Xianyuan Zhan.
\newblock {X-VLA: Soft-Prompted Transformer as Scalable Cross-Embodiment
  Vision-Language-Action Model}.
\newblock \emph{arXiv preprint arXiv:2510.10274}, 2025.

\bibitem[Zhong et~al.(2026)Zhong, He, Yang, Tian, Huang, Huang, Zhu, and
  Ma]{resvla}
Yiming Zhong, Yaoyu He, Zemin Yang, Pengfei Tian, Yifan Huang, Qingqiu Huang,
  Xinge Zhu, and Yuexin Ma.
\newblock {From Noise to Intent: Anchoring Generative VLA Policies with
  Residual Bridges}.
\newblock In \emph{Proceedings of the 43rd International Conference on Machine
  Learning}, pages 164360--164378. PMLR, 2026.

\bibitem[Zhou et~al.(2025)Zhou, Xu, Tie, Chen, Zhang, Chu, Zhou, and Sun]{pro}
Xueyang Zhou, Yangming Xu, Guiyao Tie, Yongchao Chen, Guowen Zhang, Duanfeng
  Chu, Pan Zhou, and Lichao Sun.
\newblock {LIBERO-PRO: Towards Robust and Fair Evaluation of
  Vision-Language-Action Models Beyond Memorization}.
\newblock \emph{arXiv preprint arXiv:2510.03827}, 2025.

\bibitem[Zhou et~al.(2026)Zhou, Xu, Li, Tie, Wu, Huang, Zhang, Yuan, Wu, Chu,
  Zhou, Sun, and Chen]{pro2}
Xueyang Zhou, Yangming Xu, Qianjiang Li, Guiyao Tie, Junhao Wu, Ziheng Huang,
  Guowen Zhang, Zhijie Yuan, Di Wu, Duanfeng Chu, Pan Zhou, Lichao Sun, and
  Yongchao Chen.
\newblock {LIBERO-Pro2.0: A Benchmark for Evaluating the Robustness of Embodied
  Foundation Models}.
\newblock \emph{arXiv preprint arXiv:2610.24587}, 2026.

\end{thebibliography}
}
\end{document}